\pdfoutput=1

\documentclass[11pt]{article}
\usepackage{amsmath}
\usepackage{hyperref}
\usepackage{algorithm}
\usepackage{algorithmic}
\usepackage{booktabs}
\usepackage{longtable}
\usepackage{stfloats}
\usepackage{graphicx}
\usepackage[table,xcdraw]{xcolor}
\usepackage[normalem]{ulem}
\useunder{\uline}{\ul}{}

\usepackage[]{acl}

\usepackage{times}
\usepackage{latexsym}

\usepackage[T1]{fontenc}

\usepackage[utf8]{inputenc}

\usepackage{microtype}

\usepackage{inconsolata}

\title{Measuring Bargaining Abilities of LLMs: \\ A Benchmark and A Buyer-Enhancement Method}

\author{Tian Xia\quad Zhiwei He\quad Tong Ren\quad Yibo Miao\\ {\bf Zhuosheng Zhang\quad Yang Yang\thanks{Co-corresponding Authors}\quad Rui Wang\footnotemark[1]}\\
Shanghai Jiao Tong University\\
\texttt{\{summerxt, zwhe.cs, zhangzs\}@sjtu.edu.cn,}\\
\texttt{yangyang@cs.sjtu.edu.cn, wangrui12@sjtu.edu.cn}}

\begin{document}
\maketitle
\begin{abstract}
Bargaining is an important and unique part of negotiation between humans. As LLM-driven agents learn to negotiate and act like real humans, how to evaluate agents' bargaining abilities remains an open problem.
For the first time, we formally described the Bargaining task as an asymmetric incomplete information game, defining the gains of the Buyer and Seller in multiple bargaining processes. It allows us to quantitatively assess an agent's performance in the Bargain task.
We collected a real product price dataset, \textit{AmazonHistoryPrice}, and conducted evaluations of various LLM agents' bargaining abilities\footnote{The dataset \textit{AmazonHistoryPrice} and our code are available at \href{https://github.com/TianXiaSJTU/AmazonPriceHistory}{https://github.com/TianXiaSJTU/AmazonPriceHistory}.}. We find that playing a Buyer is much harder than a Seller, and increasing model size can not effectively improve the Buyer's performance.
To address the challenge, we propose a novel approach called OG-Narrator that integrates a deterministic Offer Generator to control the price range of Buyer's offers, and an LLM Narrator to create natural language sentences for generated offers.
Experimental results show that OG-Narrator improves the buyer's deal rates from 26.67\% to 88.88\% and brings a ten times multiplication of profits on all baselines, even a model that has not been aligned.
\end{abstract}

\section{Introduction}

\begin{figure}[t]
    \centering
    \includegraphics[width=0.5\textwidth]{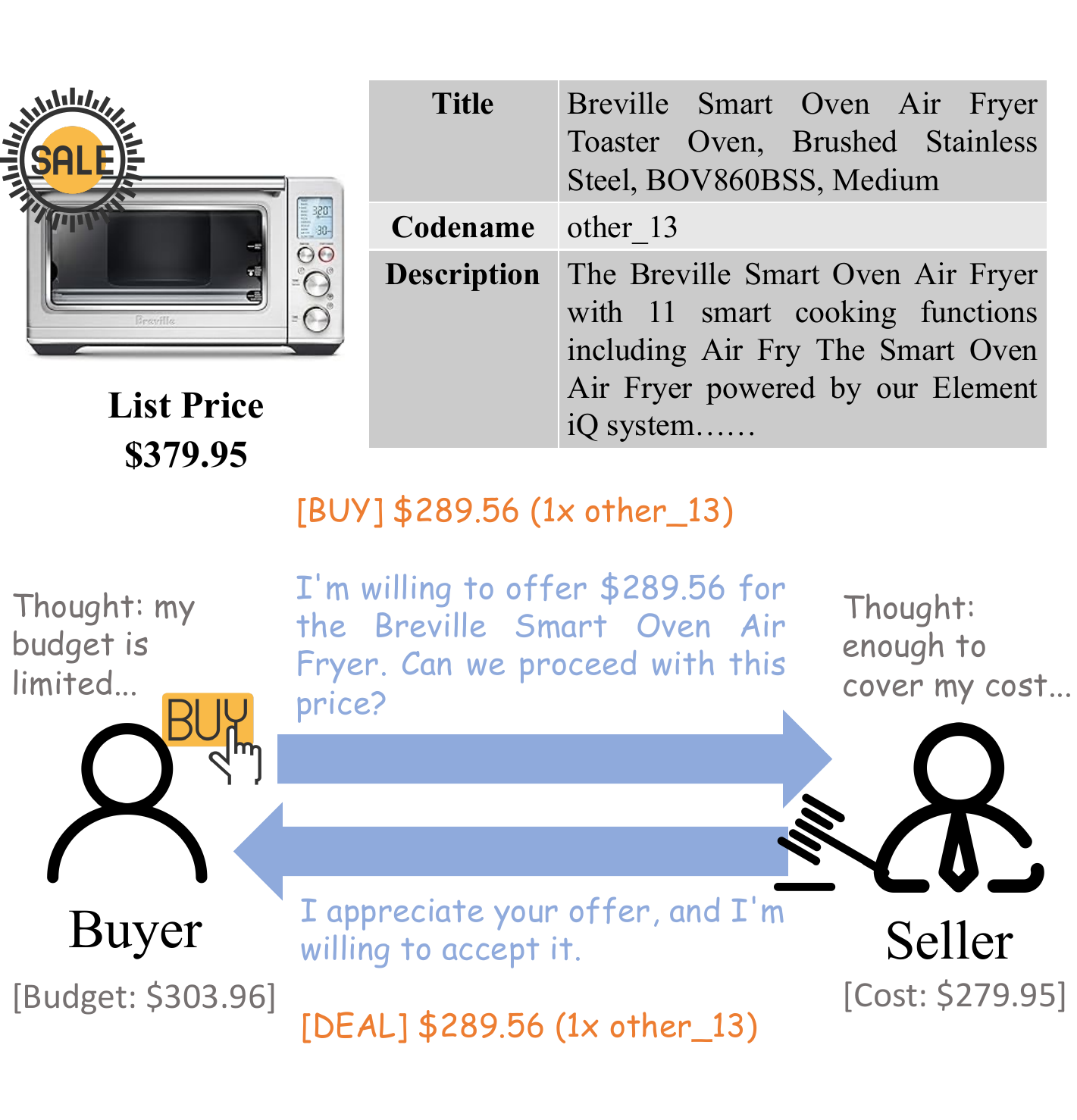}
    \vspace{-8mm}
    \caption{An example of the bargaining process. It is a simple case of two agents buying and selling an oven. Agents generate \textcolor[RGB]{80,80,80}{Thought}, \textcolor[RGB]{131,162,231}{Talk} and \textcolor[RGB]{231,151,108}{Action}, where only the \textcolor[RGB]{131,162,231}{Talk} and \textcolor[RGB]{231,151,108}{Action} are transmitted to the other party, who responds with its own \textcolor[RGB]{131,162,231}{Talk} and \textcolor[RGB]{231,151,108}{Action}. The grey text indicates the exclusive information invisible to the other party: the Buyer's \textcolor[RGB]{80,80,80}{Budget} and \textcolor[RGB]{80,80,80}{Thought} are private, as are the Seller's \textcolor[RGB]{80,80,80}{Cost} and \textcolor[RGB]{80,80,80}{Thought}.}
    \label{fig:main2}
     \vspace{-3mm}
\end{figure}

The negotiation ability of humans holds paramount significance, serving as a crucial means for humans to resolve conflicts of interest, seek mutually acceptable solutions, and facilitate the exchange of information and resources beneficial to all parties involved. This importance is underscored by various scholars such as \citet{fershtman1990importance}.

The ability to engage in bargaining is crucial for the autonomous actions of AI agents (\citealp{park2023generative}; \citealp{zhang2023igniting}; \citealp{yuan2024r}). Independent AI agents might engage in purchasing items on an e-commerce platform \citep{webshop}. In the Stanford AI Town \citep{park2023generative}, AI agents as residents of the town, might participate in multiple transactions. In many scenarios like those, unsuccessful negotiations or unreasonable bargaining could cause losses of users and unpredictable behaviors of agents in a virtual community. It is imperative to develop agents who can effectively perform price bargaining tasks to help users negotiate prices without losses and even help create a prosperous community of autonomous agents.

However, an unanswered question remains: whether the existing zero-shot capabilities \cite{kojima2023large} of the Large Language Model (LLM) are sufficiently robust to support AI agents acting as buyers or sellers, engaging in reasonable, efficient, and high-yield bargaining with other LLMs or human players. It is important to devise a method to assess the bargaining capabilities of AI agents.

Specifically, within the context of price bargaining, earlier studies \citep{craigslist,fuyao} have made preliminary explorations. However, they only leveraged a few small and domain-limited datasets to imitate the Bargaining task without clearly defining the bargaining problem.
Besides, there was a lack of analysis on how agents perform as buyers or sellers in a bargaining process.

In this paper, we made several contributions:

(i) This work formally formulated the Bargaining task for LLM agents and collected a dataset for the Bargaining task, \textit{AmazonHistoryPrice}, based on Amazon's price history, encompassing 18 categories, featuring 930 popular and real products. 

(ii) Based on our dataset, we created a benchmark to test LLMs' bargaining abilities as buyers or sellers. We tested many LLMs' performance on the benchmark, including GPT-4, ChatGPT, Llama-2, Yi, and Mistral-7B.

(iii) We proposed a simple method OG-Narrator to boost the performance of the buyer agent. We found that combining a deterministic Offer Generator and an LLM Narrator can improve the bargaining ability of a buyer agent dramatically.

\section{Dataset}

\begin{figure*}[]
    \centering
    \includegraphics[width=\linewidth]{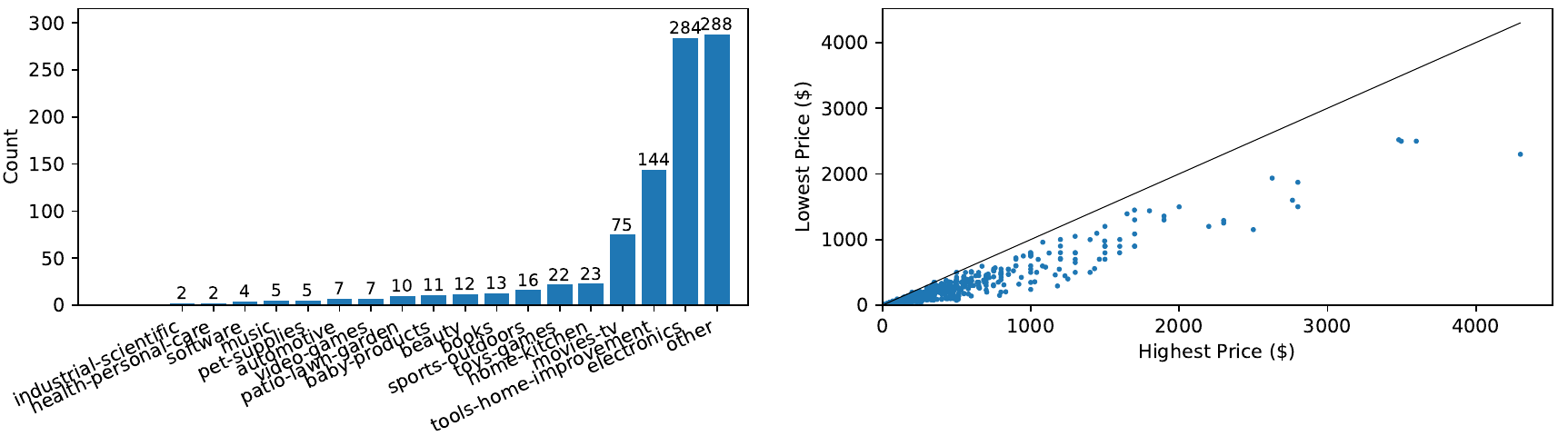}
    \caption{An overview of the diversity of our dataset \textit{AmazonHistoryPrice}. The left figure shows the categories of all items in the dataset; the right figure shows the wide range of prices. All items are from all those categories of popular products on the \href{https://camelcamelcamel.com/popular?deal=0}{camelcamelcamel} website. The imbalanced distribution of categories reflects the real-world distribution of popular items in online shopping among human users.}
    \label{fig:categories}
\end{figure*}
Current research on the communication abilities of LLM agents lacks a large real product dataset. In the work by \citet{fuyao}, a single artificially designed product (\textit{i.e.}, a balloon) was used. AucArena \citep{chen2023money} employed artificially designed products in two categories, \textit{i.e.}, cheap items and expensive items. Previous NLP research related to bargaining \citep{craigslist} introduced the \textit{CraigslistBargaining} dataset, primarily composed of different dialogues for repetitive items. Its test set comprises only 161 second-hand items.

To address this gap, we collected a dataset, \textit{AmazonHistoryPrice}, from the \href{https://camelcamelcamel.com/popular?deal=0}{camelcamelcamel} website. This dataset includes 930 Popular Products with their real prices across 18 categories: electronics, books, music, \textit{etc.}, as seen in the left figure of Figure~\ref{fig:categories}. Each product's data includes the product name, description, features, lowest and highest price, current price, list price, and an image link, as shown in Figure~\ref{fig:oven}.

\paragraph{Categories}
The commodities in this dataset represent all popular items publicly available on the website, which are those recently purchased by users\footnote{As per information from \href{https://camelcamelcamel.com/support/popular}{camelcamelcamel.com}, ``Our Popular Products show items that our users are tracking and have recently bought. By looking at the top 5-10\% most tracked products in our database and combining it with our sales reports from Amazon, we have created a page that reflects the current interests of Camel users.''}. The distribution of these items' categories mirrors the human consumers' distribution of online shopping in the real world, as seen in Figure~\ref{fig:categories}.

\paragraph{Prices}
Website records for each item include the historical lowest and highest prices, as well as the current price and corresponding dates. The price of products varies widely, ranging from 0 to 4500 USD, as illustrated in Figure~\ref{fig:categories}. The price history for some products dates back to 2009.

\paragraph{Additional Context}
Additionally, we have gathered descriptions, feature introductions, and pictures for the respective items (Figure~\ref{fig:oven}). This supplementary multi-modal information can provide AI agents with both textual and visual context.

\section{A Benchmark for Bargaining Task}
In this section, we first elaborate on the detailed definitions of the Bargaining task.
Second, we show the whole bargaining process.
Third, we describe the metrics of the Bargaining task to measure the bargaining ability of an agent in consideration of the two different kinds of scenarios.

\subsection{Task Definition}
\paragraph{Agent Bargaining Task}
The task involves two agents, the Buyer and the Seller. Both of their goals are to optimize their profits on every single session. 

Rational decision-making agents, whether Buyer or Seller, should not accept transactions resulting in negative profit. So, the Buyer would like a deal price lower than its budget, and the Seller prefers a deal price higher than the cost. However, the Buyer is unaware of the Seller's cost, and vice versa. Therefore, agents should predict the counterpart's private information based on the dialogue and combine it with their own information to decide the next move in each turn.

\paragraph{Bargaining Process}

\begin{table}[ht]
\centering
\Large
\resizebox{\columnwidth}{!}{%
\begin{tabular}{@{}llll@{}}
\toprule
Conception & Variable & Conception & Variable \\ \midrule
Session & $S$ & Action& $A$ \\
Product Info & $I$ & Budget & $B$ \\
Buyer & buyer & Cost & $C$ \\
Seller & seller & Deal Price & $D$ \\
Action Set & $S_A$ & Deal Rate & $d$ \\
Profit & $P_b, P_s$  & Normalized Profit & $P^\prime_b, P^\prime_s$ \\
Share & $\text{Share}_b, \text{Share}_s$ & Number of Sessions & $\#$ \\\bottomrule
\end{tabular}%
}
\caption{Definitions of variables in the Bargaining task.}
\label{tab:variable}
\end{table}

Our bargaining process is a variant form of the Rubinstein bargaining model \citep{rubinstein1982perfect}. To formally articulate the Bargain problem between agents, we define the relevant concepts and variables in Table~\ref{tab:variable}, and some additional explanations of concepts are in Table~\ref{tab:define}. A brief pseudo code of the process is Algorithm~\ref{alg:1}. A more vivid illustration of the process is Figure~\ref{fig:main2}.

The concept of Action is from \citet{rubinstein1982perfect}. Budget and Cost are private variables, according to \citet{gaya2021rubinstein}. The Buyer and Seller always pursue higher payoffs and avoid negative profits, based on the assumption of Individual Rationality of \citet{binmore1992noncooperative}.

Before bargaining, the Buyer needs to know the Budget, while the Seller should know the Cost. Then they take turns to talk and choose one Action from the Action Set, such as making offers and accepting offers until they have a deal or one side quits. 

\begin{algorithm}[!htb]
    \caption{Bargaining Process}
    \label{alg:1} %
    \begin{algorithmic}
        \STATE \textbf{Initialize:} Action Set $S_A$, Product Info $I$, Budget $B$, Cost $C$, Agent \text{buyer}, Agent \text{seller}, Maximum Turns $t_m$

        \STATE {$\text{buyer}\gets \text{buyer}(I,B,S_A)$}
        \STATE {$\text{seller}\gets \text{seller}(I,C,S_A)$}
        \STATE {$t\gets0$}
        \FOR{$t<t_m$}
        \STATE $A_b,\text{Talk}_b \gets \text{buyer}(I)$
        \IF {$A_b == \texttt{QUIT}$} 
        \RETURN None
        \ELSIF{$A_b == \texttt{DEAL}$} 
        \RETURN $A_b$
        \ENDIF
        \STATE $\text{seller}\gets \text{seller}(A_b,\text{Talk}_b)$
        \STATE $A_s,\text{Talk}_s \gets \text{seller}(I)$
        \IF {$A_s == \texttt{QUIT}$} 
        \RETURN None
        \ELSIF{$A_s == \texttt{DEAL}$} 
        \RETURN $A_s$
        \ENDIF
        \STATE $\text{buyer}\gets \text{buyer}(A_s,\text{Talk}_s)$
        \STATE {$t\gets t+1$}
        \ENDFOR
        \RETURN None
    \end{algorithmic}
\end{algorithm}

\paragraph{Two Scenarios}
\begin{figure}[t]
    \centering
    \includegraphics[width=0.48\textwidth]{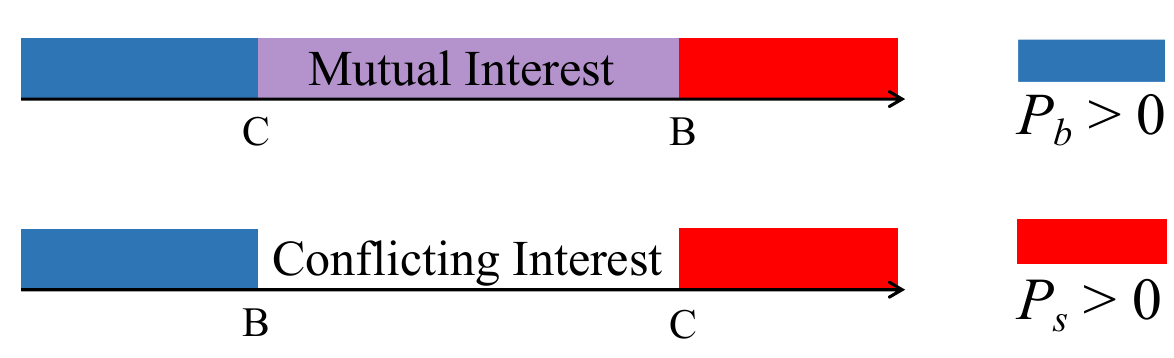}
    \caption{Two types of bargain sessions. On the axis, the \textcolor[RGB]{63,110,170}{blue} segment represents the range of $D$, which makes the Buyer's profit $P_b$ positive, while the \textcolor[RGB]{234,51,35}{red} segment signifies the range of $D$, which makes the Seller's profit $P_s$ positive. Assuming both parties are rational, the overlapping \textcolor[RGB]{175,148,202}{purple} region indicates the feasible set of the bargaining problem, \textit{i.e.}, the set of all possible deal prices for both sides. In other regions, one of them should always reject the price.}
    \label{fig:interest}
\end{figure}

We separately define the Buyer's profit and the Seller's profit with Budget, Cost, and Deal Price:
\begin{equation}\label{eq:profit}
    P_b = B-D,\quad
    P_s = D-C.
\end{equation}

With different Budgets and Costs, sessions can be divided into two types: \textbf{Mutual Interest (MI)} and \textbf{Conflicting Interest (CI)}, as depicted in Figure~\ref{fig:interest}. Mutual Interest includes the set of all possible agreements, while Conflicting Interest represents the possibility that two agents never reach an agreement \citep{binmore1986nash}.

In our Bargaining task, according to \citet{rubinstein1982perfect}, Rubinstein's model sets the utility functions of the Buyer and Seller as
\begin{equation}\label{eq:utility}
\begin{split}
u_b &= \frac{B-D}{B-C} = \frac{P_b}{B-C},\\
u_s &= \frac{D-C}{B-C} = \frac{P_s}{B-C}.
\end{split}
\end{equation}

However, Rubinstein's model only includes the situation in which it is possible to deal because of mutual interest \citep{binmore1986nash}. When in MI scenarios, $B > C$, one side's utility $u > 0$ if and only if its profit $P>0$. However, in CI scenarios, $B \le C$, according to Equation~(\ref{eq:utility}), one side's utility $u > 0$ if and only if $P<0$, which is inconsistent and counter-intuitive.

\paragraph{Metrics}
We define the \textbf{Normalized Profits (NP)} for the Buyer and Seller as:
\begin{equation}\label{eq:ratio}
P_b^{\prime}=\frac{B-D}{|B-C|},\ 
P_s^{\prime}=\frac{D-C}{|B-C|}.
\end{equation}
In sessions that make no deal, $D$ does not exist, so the NPs of both sides are 0.
To prevent division by zero errors, in the case of $B=C$, we set $B=C-\sigma$ ($\sigma$ is a small offset). NP satisfies the constraints of Rubinstein's model and can be compared across two types of scenarios.

To evaluate models on multiple sessions, we define \textbf{Sum of Normalized Profits (SNP)} for the Buyer and Seller as follows:
\begin{equation}\label{eq:SNP}
    \text{SNP}_b = \sum\limits_{i=1}^{N}{P_b^{\prime (i)}},\quad
    \text{SNP}_s = \sum\limits_{i=1}^{N}{P_s^{\prime (i)}},
\end{equation}
where $N$ can be the number of all 930 sessions in our dataset, or the number of MI or CI sessions. Higher SNP means better bargaining ability.

Also, we define the valid session as a session that ends correctly with the specific output formats we set. All possible normalized profits on the table from all valid sessions are equal to the number of valid MI deals minus the number of valid CI deals, derived from Equation~\ref{eq:ratio} and Equation~\ref{eq:SNP}:
\begin{equation}\label{eq:allNP}
    \text{SNP}_b + \text{SNP}_s = \#_{\text{MI}}* d_{\text{MI}} - \#_{\text{CI}}* d_{\text{CI}}
\end{equation}

Moreover, to measure the division of all possible profit on the table from all valid sessions, we define the \textbf{Share} of the Buyer and Seller as:
\begin{equation}\label{eq:share}
    \text{Share}_b = \frac{\text{SNP}_b}{\text{SNP}_b+\text{SNP}_s},\ 
    \text{Share}_s = \frac{\text{SNP}_s}{\text{SNP}_b+\text{SNP}_s}.
\end{equation}
If the Buyer's Share is higher than the Seller's Share, $\text{Share}_b > 50\%$, the Buyer makes more profits than the Seller in the whole evaluation. The Share is undefined when $\text{SNP}_b+\text{SNP}_s \leq 0$. 

\section{Experiments}
In this section, we report the implementation details and the benchmark performances of many well-known LLMs against ChatGPT in the Bargaining task on our dataset. LLMs are listed in Appendix~\ref{a:llms}.

\subsection{Implementation Details}
We use vLLM \cite{vllm} to run all models on 2 Nvidia H800 GPUs with the same seed. Evaluation of a 34B model over all 930 products in our dataset takes roughly 1 hour on a single H800. We use OpenAI API gpt-3.5-turbo-1106 as ChatGPT and gpt-4-0125-preview as GPT-4.

We set all temperatures to 0. Due to OpenAI's closed-source non-deterministic implementation, small changes may still occur in the reproduction process. Our code supports the Bargaining of two locally deployed LLMs for a more deterministic reproduction.

We chose ChatGPT as both Buyer and Seller against all LLMs, because ChatGPT is one of the best-performing LLM agents, compared to other open-source models. 1) ChatGPT has a strong enough instruction-following ability and rarely produces format errors. It correctly completed 94.3\% of all sessions; 2) It performs more closely to humans compared to other models because they rarely accept irrational deals in CI scenarios, as seen in Appendix~\ref{a:humanAndChatGPT}.

To test the out-of-the-box capabilities of LLM, we employed LLM itself as the agent without incorporating any additional modules, such as memory or backtracking. We adopted the Chain of Thought (CoT) approach \cite{wei2023chainofthought}, informed the model about the instructions of the Bargaining task and the specified format for dialogue generation, producing Thought, Talk, and Action in each interaction. Prompts are in Appendix~\ref{a:prompts}.

\subsection{Variable Initialization}

\begin{figure}[t]
    \centering
    \includegraphics[width=0.45\textwidth]{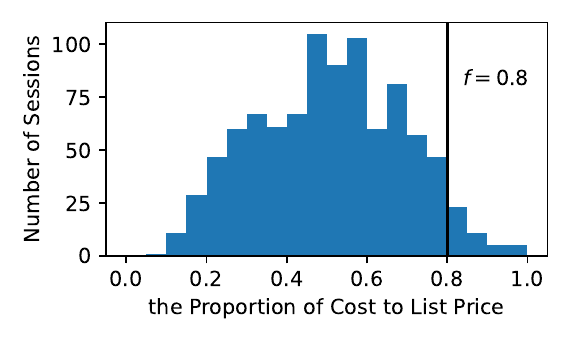}
    \caption{The distribution of the proportions of Cost to List Price in our dataset. MI sessions' proportions are lower than $f$, while those higher are CI sessions.} %
    \label{fig:priceratio}
\end{figure}

\paragraph{Product Info}
The Product Info includes the product title, description, codename, and List Price, as illustrated in Figure~\ref{fig:main2}. We use the historical highest price as the List Price. List Price cannot be set to current price because the longer a product has been on the market, the lower the price becomes. Current price is therefore related to the product's listing time, which would introduce variables unrelated to the product itself.

We only include one single product in each session, due to the complexity of bargaining on multiple items.

\paragraph{Cost}
We cannot directly obtain the actual cost of products sold on Amazon. However, over a sufficiently long historical period, Amazon's lowest discounted price may approach the cost price. In this dataset, the lowest price is on average 51.08\% of the highest price.
Since the historical price data for the products dates back to 2009, our experiment sets the Cost $C$ for each product as the historical lowest price observed in the dataset.

\begin{table*}[!htbp]
\begin{minipage}[b]{\linewidth}
\centering
\resizebox{\textwidth}{!}{%
\begin{tabular}{@{}lrrrlrrrlrrr@{}}
\toprule
 & \multicolumn{3}{c}{ALL} & \textbf{} & \multicolumn{3}{c}{MI} & \textbf{} & \multicolumn{3}{c}{CI} \\ \cmidrule(lr){2-4} \cmidrule(lr){6-8} \cmidrule(l){10-12} 
Buyer & \multicolumn{1}{c}{\#} & \multicolumn{1}{c}{$\text{SNP}_b$} & \multicolumn{1}{c}{$\text{Share}_b$} & \multicolumn{1}{c}{} & \multicolumn{1}{c}{\#} & \multicolumn{1}{c}{Deal rate} & \multicolumn{1}{c}{$\text{SNP}_b$} & \multicolumn{1}{c}{\textbf{}} & \multicolumn{1}{c}{\#} & \multicolumn{1}{c}{Deal rate} & \multicolumn{1}{c}{$\text{SNP}_b$} \\ \midrule
GPT-4 & 851 & \textbf{-33.81} & \textbf{-11.27\%} &  & 807 & 37.55\% & \textbf{-23.46} &  & 44 & 6.82\% & -10.35 \\
Mixtral-8x7B-Instruct & 505 & -63.19 & -43.28\% &  & 475 & 31.79\% & -59.66 &  & 30 & 16.67\% & -3.53 \\
Mistral-7B-Instruct & 786 & -89.17 & -27.87\% &  & 748 & 44.92\% & -77.32 &  & 38 & 42.11\% & -11.85 \\
Yi-6B-Chat & 561 & -122.94 & -87.19\% &  & 532 & 27.44\% & -116.16 &  & 29 & 17.24\% & -6.78 \\
Yi-34B-Chat & 761 & -129.76 & -54.52\% &  & 722 & 33.66\% & -111.20 &  & 39 & 12.82\% & -18.56 \\
Qwen-14B-Chat & 562 & -159.21 & -70.13\% &  & 529 & 44.61\% & -121.24 &  & 33 & 27.27\% & -37.96 \\
ChatGPT & \textbf{877} & -164.52 & -59.61\% &  & 835 & 34.01\% & -157.73 &  & 42 & 19.05\% & -6.80 \\
Baichuan2-13B-Chat & 510 & -216.67 & -152.59\% &  & 484 & 30.58\% & -237.84 &  & 26 & 23.08\% & \textbf{21.17} \\
ChatGLM3-6B & 546 & -261.91 & -137.13\% &  & 516 & 38.57\% & -219.25 &  & 30 & 26.67\% & -42.66 \\
Llama-2-7b-chat & 612 & -288.59 & -113.17\% &  & 576 & 45.83\% & -279.60 &  & 36 & 25.00\% & -8.99 \\
Llama-2-13b-chat & 720 & -305.53 & -82.35\% &  & 682 & 56.30\% & -270.11 &  & 38 & 34.21\% & -35.43 \\
Llama-2-70b-chat & 660 & -361.26 & -127.20\% &  & 625 & 47.36\% & -335.93 &  & 35 & 34.29\% & -25.33 \\
Baichuan2-7B-Chat & 653 & -603.67 & -199.23\% &  & 623 & 50.40\% & -567.11 &  & 30 & 36.67\% & -36.57 \\
Qwen-7B-Chat & 647 & -753.16 & -201.92\% &  & 615 & 62.44\% & -692.12 &  & 32 & 34.38\% & -61.04 \\ \bottomrule
\end{tabular}%
}
\end{minipage}

\vspace{0.5em}

\begin{minipage}[b]{\linewidth}
\centering
\resizebox{\textwidth}{!}{%
\begin{tabular}{@{}lrrrlrrrlrrr@{}}
\toprule
 & \multicolumn{3}{c}{ALL} & \textbf{} & \multicolumn{3}{c}{MI} & \textbf{} & \multicolumn{3}{c}{CI} \\ \cmidrule(lr){2-4} \cmidrule(lr){6-8} \cmidrule(l){10-12} 
Seller & \multicolumn{1}{c}{\#} & \multicolumn{1}{c}{$\text{SNP}_s$} & \multicolumn{1}{c}{$\text{Share}_s$} & \multicolumn{1}{c}{} & \multicolumn{1}{c}{\#} & \multicolumn{1}{c}{Deal rate} & \multicolumn{1}{c}{$\text{SNP}_s$} & \multicolumn{1}{c}{\textbf{}} & \multicolumn{1}{c}{\#} & \multicolumn{1}{c}{Deal rate} & \multicolumn{1}{c}{$\text{SNP}_s$} \\ \midrule
GPT-4 & \textbf{930} & \textbf{1178.15} & 137.31\% &  & 886 & 98.87\% & \textbf{1153.13} &  & 44 & 40.91\% & \textbf{25.02} \\
Yi-34B-Chat & 899 & 579.33 & 86.60\% &  & 859 & 80.79\% & 590.90 &  & 40 & 62.50\% & -11.56 \\
Mistral-7B-Instruct & 830 & 526.50 & 78.58\% &  & 791 & 89.25\% & 569.45 &  & 39 & 92.31\% & -42.95 \\
Mixtral-8x7B-Instruct & 600 & 483.99 & 110.75\% &  & 574 & 79.09\% & 496.82 &  & 26 & 65.38\% & -12.83 \\
ChatGPT & 877 & 440.52 & \textbf{159.61\%} &  & 835 & 34.01\% & 441.73 &  & 42 & 19.05\% & -1.20 \\
Llama-2-70b-chat & 837 & 415.28 & 64.38\% &  & 797 & 84.82\% & 451.93 &  & 40 & 77.50\% & -36.65 \\
Qwen-14B-Chat & 795 & 393.16 & 71.10\% &  & 759 & 75.89\% & 421.24 &  & 36 & 63.89\% & -28.08 \\
Llama-2-13b-chat & 727 & 308.21 & 62.90\% &  & 693 & 74.31\% & 334.53 &  & 34 & 73.53\% & -26.32 \\
Qwen-7B-Chat & 793 & 92.86 & 37.14\% &  & 752 & 35.24\% & 114.63 &  & 41 & 36.59\% & -21.77 \\
ChatGLM3-6B & 701 & 91.10 & 17.83\% &  & 675 & 78.37\% & 160.94 &  & 26 & 69.23\% & -69.84 \\
Llama-2-7b-chat & 496 & 49.54 & 16.24\% &  & 471 & 67.52\% & 71.82 &  & 25 & 52.00\% & -22.28 \\
Baichuan2-7B-Chat & 762 & 38.29 & 6.36\% &  & 728 & 86.40\% & 153.60 &  & 34 & 79.41\% & -115.31 \\
Yi-6B-Chat & 64 & 14.14 & 27.72\% &  & 60 & 91.67\% & 31.29 &  & 4 & 100.00\% & -17.15 \\
Baichuan2-13B-Chat & 741 & -211.92 & -48.16\% &  & 701 & 66.48\% & -164.63 &  & 40 & 65.00\% & -47.29 \\ \bottomrule
\end{tabular}%
}
\end{minipage}
\caption{The performances of various LLMs playing the Buyer and Seller in the Bargaining task, sorted in descending order of the SNP in ALL. SNP is used for comparisons across models, while Share is used for comparisons between that model and its counterpart ChatGPT.
 (\textbf{SNP} is the Sum of Normalized Profits, defined in Equation~\ref{eq:SNP}. \textbf{Share} is a party's share of the overall profit from all valid sessions, defined in Equation~\ref{eq:share}. \# in MI/CI means the number of valid MI/CI sessions. The deal rate $d$ in MI/CI means the proportion of deals to valid MI/CI sessions.)}
\label{tab:bench}
\end{table*}

\paragraph{Budget}
To better control variables and avoid manually setting specific budget values for each product, we use budget factor $f$ to set each product's budget $B=fL$. By adjusting $f$, we can easily change the proportion of MI sessions in the Bargaining task. To set a moderate level of difficulty for the Bargaining task, we set $f=0.8$ to control the proportion of CI sessions to only 5\%, because the model's behavior may be more chaotic and random when rational deals cannot be made in CI sessions.

If we set $f$ lower, the profit margin would be too small, and the CI proportion would be too high. If we set $f$ higher, the profit margin would be large, making the task too simple and lacking in the assessment of real bargaining abilities. The results of different $f$ are shown in Appendix~\ref{a:factors}.

\paragraph{Action}
\citet{heddaya2023language} annotated Bargaining Acts in text and investigated the impact of acts such as New Offer and Push. Inspired by this approach, we designed five core actions to represent the agents' intentions in the bargaining process. These actions are BUY, SELL, DEAL, REJECT, and QUIT, with the specific meanings, as seen in Table~\ref{tab:action} in Appendix~\ref{a:Action}. AI agents often need to call functions to make offers and determine transaction outcomes. Our implementation is letting agents output action strings in a designated format.

\subsection{Benchmark Results}
We report the performance of various LLMs as Buyers and Sellers separately in Table~\ref{tab:bench}.

\paragraph{Buyer Performance}
First of all, almost all models we tested have negative SNP in our benchmark, which means that, on average, they make a negative profit every time they close a deal. This indicates that they cannot rigorously obey the basic rule of bargaining given in the prompt in Table \ref{tab:buyerprompts}: ``\textit{You can only buy things that cost less than your budget; otherwise, you should quit negotiating.}''.

Although all Buyers' SNPs are negative, GPT-4 has the least profit loss among all models due to its best performance in MI sessions. ChatGPT has the most valid sessions, outperforming GPT-4. Among all open-source models, Mistral-7B and Mixtral-8x7B show the best performances. Mistral-7B has more valid sessions and a higher Share when competing with ChatGPT, next to GPT-4's Share, while Mixtral-8x7B has a higher SNP in both MI and CI sessions.

\begin{figure}
    \centering
    \includegraphics[width=\linewidth]{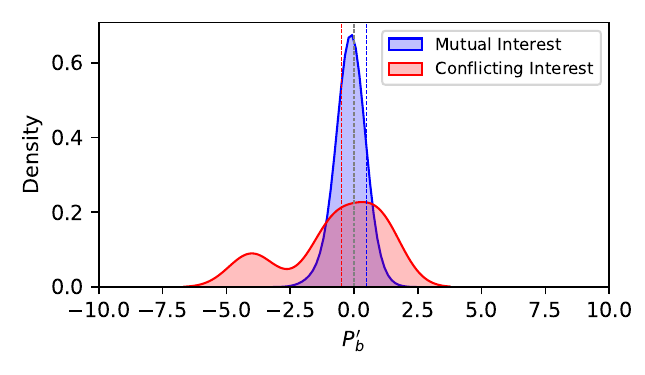}
    \caption{The distribution of Buyer's normalized profits of all sessions, when Mixtral-8x7B plays Buyer and ChatGPT plays Seller. The average of normalized profits is slightly below zero. \textcolor{red}{Red} dashed line is -0.5, and the \textcolor{blue}{blue} line is 0.5. They separate those Buyers who gain more than Seller and those who do not, in CI and MI.}
    \label{fig:buyerkde}
\end{figure}

Negative profits in MI indicate the gap between the Buyer and Seller in the Bargaining task. Examining the profit distribution of Mixtral-8x7B in Figure~\ref{fig:buyerkde}, even the second-best Buyer model has more than half of its sessions with negative profits.

An exception to the negative SNP trend is Baichuan2-13B in CI sessions. We examined all 6 CI deals and demonstrated the only case with a positive profit in Table~\ref{tab:case1}. In this case, the ChatGPT Seller mistakes the Buyer's hallucination Talk for the real list price and accepts a deal far below the real cost.

\paragraph{Seller Performance}
The SNPs in ALL sessions of all models are positive, except for Baichuan2-13B. GPT-4 has achieved the best performance with the highest SNP in both MI and CI sessions as the Seller. GPT-4 also has a 100\% valid rate for all 930 sessions.

ChatGPT has the highest Seller's Share 159.61\%, GPT-4 has 137.31\% and Mixtral-8x7B-Instruct has 110.75\%. Those Shares greater than 100\% mean that these three Sellers are exploiting the Buyers and making excess profits.




\section{Discussion}
In this section, we discuss our findings from the benchmarks for Buyers and Sellers.

\paragraph{Playing Buyer is more difficult than playing Seller.}
As seen in Table~\ref{tab:bench}, ChatGPT as the Seller beats all Buyers, even ChatGPT itself. But when playing Buyer it loses to GPT-4 and Mixtral-8x7B-Instruct. Even when the abilities of agents are relatively equal, bargaining as a Buyer is more challenging than bargaining as a Seller, which is also observed in \citet{fuyao}.

The difficulties include 1) the Buyer should correctly understand the Buyer role along with the concept of Budget and realize that the deal price should not exceed the Budget to avoid loss; 2) the Buyer needs to actively persuade the conservative Seller who prefers to use REJECT actions, as seen in the dialogue from Table~\ref{tab:case2}.

\paragraph{For Buyer, different-sized models from the same family have close performance.}
Qwen-7B, Qwen-13B, and Baichuan2-13B seem not to fully understand the prompt for the Buyer. They have a high probability, which is 99\% for Qwen-7B, 95\% for Qwen-14B, and 69\% for Baichuan2-13B, of perceiving themselves as the Seller from the beginning of the conversation, as seen in Table~\ref{tab:case3}. In many cases, they still generate at least one valid BUY Action as the Buyer in a whole conversation, so we did not remove these cases from the benchmark results. However, this phenomenon did introduce more unpredictability to the deal price.

Apart from Qwen-7B, Qwen-13B, and Baichuan2-13B, we found that different-sized models of the same series, including Llama-2, Yi, and Mistral, have similar performances as a Buyer in Table~\ref{tab:bench}. Considering nearly identical decoder architecture \cite{llama2, ai2024yi, yang2023baichuan}, this indicates that it is not the model size but the training corpus and the method of training and aligning that affect the Buyer’s bargaining performance.

\paragraph{Most models start from a high bid.}
In most of the cases we studied, the Buyer tends to start with a slightly lower price than the given Budget. It heavily affects the final deal price if a deal could be made. As demonstrated in Table~\ref{tab:case4}, the Buyer's starting offer was \$30, only 2 dollars lower than its Budget, which means the room for the Buyer's profit was limited to 2 dollars.

We define \textbf{First Bid Ratio (FBR)} as the ratio of the Buyer's first bid's price to the Buyer's Budget and then calculate the average First Bid Ratio in all valid sessions for each Buyer. As seen in Table~\ref{tab:humanBuyer} and Table~\ref{tab:firstBidRatio}, compared to LLMs, Human Buyer has a significantly lower FBR and much higher SNP and Share against ChatGPT. The Buyer should start with a very low price and escalate the offer price. To achieve this goal, we proposed a new buyer-enhancement method OG-Narrator to decrease the FBR and demonstrated its effectiveness on all kinds of models by experiments on the Buyer benchmark.

\begin{table}[]
\centering
\resizebox{\columnwidth}{!}{%
\begin{tabular}{@{}lrrrr@{}}
\toprule
Buyer & \multicolumn{1}{c}{\#} & \multicolumn{1}{c}{Avg.FBR} & \multicolumn{1}{c}{$\text{SNP}_b$} & \multicolumn{1}{c}{$\text{Share}_b$} \\ \midrule
GPT-4                 & 851 & 0.90 & -33.81  & -11.27\%  \\
Mixtral-8x7B-Instruct & 505 & 0.87 & -63.19  & -43.28\%  \\
Mistral-7B-Instruct   & 786 & 0.86 & -89.17  & -27.87\%  \\
Yi-6B-Chat            & 561 & 0.85 & -122.94 & -87.19\%  \\
Yi-34B-Chat           & 761 & 0.97 & -129.76 & -54.52\%  \\
Qwen-14B-Chat         & 562 & 1.00 & -159.21 & -70.13\%  \\
ChatGPT               & 877 & 0.91 & -164.52 & -59.61\%  \\
Baichuan2-13B-Chat    & 510 & 0.81 & -216.67 & -152.59\% \\
ChatGLM3              & 546 & 0.86 & -261.91 & -137.13\% \\
Llama-2-7b-chat       & 612 & 0.81 & -288.59 & -113.17\% \\
Llama-2-13b-chat      & 720 & 0.94 & -305.53 & -82.35\%  \\
Llama-2-70b-chat      & 660 & 0.92 & -361.26 & -127.20\% \\
Baichuan2-7B-Chat     & 653 & 0.96 & -603.67 & -199.23\% \\
Qwen-7B-Chat          & 647 & 1.00 & -753.16 & -201.92\% \\ \bottomrule
\end{tabular}%
}
\caption{The average ratios of Buyer's first bid compared to Buyer's budget in all valid sessions. First Bid Ratio (FBR) = Buyer's First Bid / Budget. Seller: gpt-3.5-turbo-1106.}
\label{tab:firstBidRatio}
\end{table}

\paragraph{For Seller, model size matters.}
In contrast to the Buyer, the Seller's SNP is highly related to the valid rate. A high valid rate requires good instruction in the following capabilities. As observed in Table~\ref{tab:bench}, we found that the Seller model with a bigger size of parameters in the same series tends to have a higher valid rate and higher deal rate. For instance, Llama-2-70b is the best among all Llama-2 models and Yi-34B is also the best among Yi models. Also, all the worst models' sizes are about 7B. 

The only exceptions are Mixtral-8x7B and Baichuan2-13B-Chat. Mixtral-8x7B applies the Mixture of Experts (MoE) mechanism on the same size of Transformer as Mistral-7B \citep{jiang2024mixtral, jiang2023mistral}. Its extra parameters of expert layers do not improve its valid rate and SNP, but increase the Share as a Seller. Baichuan2-13B-Chat has a lower valid rate, a lower deal rate, and a lower SNP than Baichuan2-7B-Chat. The reason could be its positional embedding ALiBi, which is a special kind of relative attention \cite{alibi}, that makes the model ignore the critical information for Seller from the long dialogue, such as the cases of Table~\ref{tab:case5} and Table~\ref{tab:case6}.

\begin{table*}[!hbtp]
\centering
\resizebox{\textwidth}{!}{%
\begin{tabular}{@{}lrrrrlrrrlrrr@{}}
\toprule
                                 & \multicolumn{4}{c}{ALL} &  & \multicolumn{3}{c}{MI} &  & \multicolumn{3}{c}{CI} \\ \cmidrule(lr){2-5} \cmidrule(lr){7-9} \cmidrule(l){11-13} 
Buyer &
  \multicolumn{1}{c}{\#} &
  \multicolumn{1}{c}{Avg.FBR} &
  \multicolumn{1}{c}{{$\text{SNP}_b$}} &
  \multicolumn{1}{c}{{$\text{Share}_b$}} &
   &
  \multicolumn{1}{c}{\#} &
  \multicolumn{1}{c}{Deal rate} &
  \multicolumn{1}{c}{{$\text{SNP}_b$}} &
   &
  \multicolumn{1}{c}{\#} &
  \multicolumn{1}{c}{Deal rate} &
  \multicolumn{1}{c}{{$\text{SNP}_b$}} \\ \midrule
\rowcolor[HTML]{C0C0C0} 
Phi-2                            & 47  & 0.42 & -1.58   & -31.56\%  &  & 44     & 13.64\%    & -0.38     &  & 3      & 33.33\%    & -1.20     \\
\multicolumn{1}{r}{+OG-Narrator} & 890 & 0.50 & 2015.98 & 284.74\%  &  & 847    & 88.08\%    & 1868.69   &  & 43     & 88.37\%    & 147.28    \\
\rowcolor[HTML]{C0C0C0} 
Yi-6B-Chat                       & 561 & 0.85 & -122.94 & -87.19\%  &  & 532    & 27.44\%    & -116.16   &  & 29     & 17.24\%    & -6.78     \\
\multicolumn{1}{r}{+OG-Narrator} & 915 & 0.50 & 1798.67 & 261.43\%  &  & 872    & 82.68\%    & 1709.52   &  & 43     & 76.74\%    & 89.16     \\
\rowcolor[HTML]{C0C0C0} 
Yi-34B-Chat                      & 761 & 0.97 & -129.76 & -54.52\%  &  & 722    & 33.66\%    & -111.20   &  & 39     & 12.82\%    & -18.56    \\
\multicolumn{1}{r}{+OG-Narrator} & 899 & 0.50 & 1650.37 & 225.15\%  &  & 855    & 89.59\%    & 1505.19   &  & 44     & 75.00\%    & 145.18    \\
\rowcolor[HTML]{C0C0C0} 
Llama-2-70b-chat                 & 660 & 0.92 & -361.26 & -127.20\% &  & 625    & 47.36\%    & -335.93   &  & 35     & 34.29\%    & -25.33    \\
\multicolumn{1}{r}{+OG-Narrator} & 790 & 0.50 & 1395.60 & 226.56\%  &  & 754    & 85.54\%    & 1276.39   &  & 36     & 80.56\%    & 119.21    \\
\rowcolor[HTML]{C0C0C0} 
Baichuan2-13B-Chat               & 510 & 0.81 & -216.67 & -152.59\% &  & 484    & 30.58\%    & -237.84   &  & 26     & 23.08\%    & 21.17     \\
\multicolumn{1}{r}{+OG-Narrator} & 863 & 0.50 & 1316.36 & 214.39\%  &  & 821    & 78.44\%    & 1210.67   &  & 42     & 71.43\%    & 105.69    \\
\rowcolor[HTML]{C0C0C0} 
Mixtral-8x7B-Instruct            & 505 & 0.87 & -63.19  & -43.28\%  &  & 475    & 31.79\%    & -59.66    &  & 30     & 16.67\%    & -3.53     \\
\multicolumn{1}{r}{+OG-Narrator} & 846 & 0.50 & 1048.65 & 179.87\%  &  & 808    & 75.99\%    & 937.11    &  & 38     & 81.58\%    & 111.55    \\
\rowcolor[HTML]{C0C0C0} 
Qwen-7B-Chat                     & 647 & 1.00 & -753.16 & -201.92\% &  & 615    & 62.44\%    & -692.12   &  & 32     & 34.38\%    & -61.04    \\
\multicolumn{1}{r}{+OG-Narrator} & 904 & 0.50 & 975.33  & 141.15\%  &  & 863    & 83.20\%    & 899.35    &  & 41     & 65.85\%    & 75.98     \\ \bottomrule
\end{tabular}%
}
\caption{The performances of some models using OG-Narrator compared to the original Buyer benchmarks. First Bid Ratio (FBR) = Buyer's First Bid / Budget.}
\label{tab:botbench}
\end{table*}

\section{A Simple Bargaining Method: OG-Narrator}
\label{sec:og}
In this section, we propose a new method, OG-Narrator, to enhance an LLM agent's bargaining performance.

We believe that LLMs using CoT do not fully exploit the potential of the Buyer in the Bargaining task. And also, currently the Buyer's weak performance limits the fair assessment of the ChatGPT Seller. It is unknown if the ChatGPT Seller's performance remains good when their opponents are improved. So we design a method OG-Narrator in Section~\ref{sec:og} to improve the Buyer and re-evaluate the ChatGPT Seller as well.
\subsection{Method}

\begin{figure}[bt]
    \centering
    \includegraphics[width=\columnwidth]{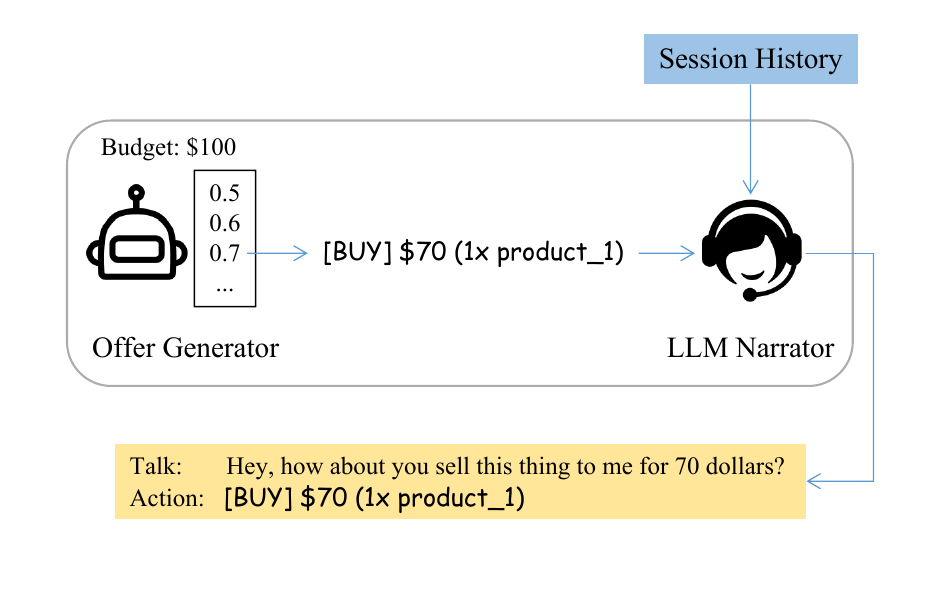}
    \caption{An overview of OG-Narrator. In each turn, the Offer Generator generates an offer (Action) to the LLM Narrator, then the LLM narrates the offer in natural language based on session history, and finally outputs Talk and Action.}
    \label{fig:OGMethod}
\end{figure}

\begin{figure}[bt]
    \centering
    \includegraphics[width=\linewidth]{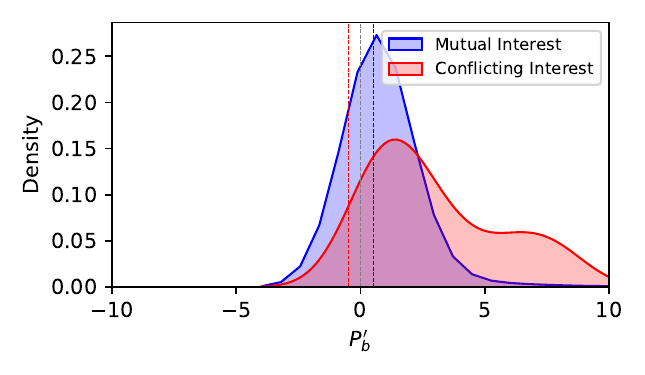}
    \caption{The new distribution of Buyer's normalized profits of all sessions, when Mixtral-8x7B using OG-Narrator plays Buyer and ChatGPT plays Seller. The average of normalized profits increases.}
    \label{fig:botkde}
\end{figure}

\citet{mannekote2023neural} proposed a pipeline-based architecture for a dialogue system. Inspired by this work, we have designed a similar method, named OG-Narrator, to enhance an LLM agent for better performance as a Buyer, by decreasing the price of the first bid.

As seen in Figure~\ref{fig:OGMethod}, OG-Narrator deploys an \textbf{Offer Generator} (OG) to help generate prices for the Buyer's offers and let the LLM generate natural language sentences based on given offers. The deterministic Offer Generator first produces factors escalating from 0.5 to 1 using linear interpolation, and then it multiplies the budget and one of the factors to get a new offer price $p = (0.5 + 0.5\frac{t}{t_{m}})B$ in each turn $t$. After receiving the Seller's offer, if the offer's price is still higher than $p$, the Buyer proposes a new BUY Action $A_b$ with the price $p$; otherwise, the Buyer chooses to deal.

Moreover, OG-Narrator employs the original agent as an \textbf{LLM Narrator}, to generate Talk for Buyer based on an Action and the session history, $\text{Talk}_b=\text{LLM}(I,A_b)$. Consequently, OG-Narrator frees the LLM from generating Actions and lets LLM only focus on natural language. 

\subsection{Experiments}
We ran the Buyer benchmark on selected models using OG-Narrator and compared it to the original benchmark in Table~\ref{tab:botbench}.

To test the model that has not been aligned to complex chat tasks, we also added a model Phi-2 to the test, which is a 2.7B model specialized for basic Python coding \cite{phi1} and has not been fine-tuned through reinforcement learning from human feedback~\cite{he2024improving,ouyang2022training}, according to the home \href{https://huggingface.co/microsoft/phi-2}{page}.

\subsection{Results}
As seen in Table~\ref{tab:botbench}, OG-Narrator successfully controls each Buyer's FBR to 0.5 as we set. Compared to the original benchmark, our method significantly improves the performance of all models, even an unaligned model.

\paragraph{OG-Narrator vastly improves the valid rate, deal rate, SNP, and Share of all models.}
As shown in Figure~\ref{fig:botkde}, the SNP remarkably increases, especially in CI sessions, compared to Figure~\ref{fig:buyerkde}. This demonstrates that the OG-Narrator method alleviates the difficulty of bargaining as a Buyer by decoupling the offer generation strategy.

\paragraph{OG-Narrator enables the unaligned model to bargain as a Buyer.}
Among all models, only Phi-2 has not been fine-tuned for chat nor aligned through RLHF. As seen in Table~\ref{tab:botbench}, its valid rate and deal rate are so low that unaligned Phi-2 cannot effectively play the Buyer role in the Bargaining task. However, applying OG-Narrator to Phi-2 dramatically increases the valid rate by 19 times and the deal rate by 117 times, making the Phi-2 agent much more reliable than before, even though it is still not aligned.

\paragraph{The ChatGPT Seller is vulnerable.}
OG-Narrator uses a simple linear function as the strategy to generate offers for the Buyer agent. This straightforward strategy helps all LLMs, even weaker ones, achieve over 100\% Share and exploit the ChatGPT Seller in Table~\ref{tab:botbench}, in contrast to the negative Shares in the original benchmark  Table~\ref{tab:bench}. It implies that the ChatGPT Seller is vulnerable and not robust enough when facing a Buyer enhanced by the simple strategy tool OG-Narrator.

\section{Related Work}
In this section, we listed related works from the perspectives of AI agents and Bargaining.

\paragraph{AI Agents}
The memory, planning, reasoning, and communication capabilities possessed by large-scale LLMs bring hope for the development of Autonomous AI agents (\citealp{autogpt}; \citealp{park2023generative}; \citealp{liang2023encouraging}; \citealp{liang2023leveraging}; \citealp{ai2024cognition}; \citealp{10.1162/tacl_a_00642}).
Generative agents \citep{park2023generative} have created a town filled with independent agents, each playing different roles and autonomously engaging in social interactions with other agents. 

\paragraph{Bargaining}
Previous work \cite{craigslist} proposed a small dataset of bargaining dialogues on second-hand items. However, both the Buyer and Seller lack reasonable mental expectations for the prices of second-hand items. Recent work \cite{fuyao} evaluated the Buyer and Seller based on one item's final deal price only. However, their method can not evaluate the bargaining performance on multiple sessions. More details and other works are discussed in Appendix~\ref{a:bargaining}.

\section{Conclusion}
To the best of our knowledge, we formally described the Bargaining task for the first time, defining the evaluation metrics for the Buyer and Seller to quantitatively assess agent performance.

We collected a large-scale real product price dataset, \textit{AmazonHistoryPrice}, for the bargaining task. Using this dataset, we evaluated the bargaining abilities of various LLMs. Our findings indicate that playing the Buyer is more difficult than playing the Seller. We found that larger models lead to better Seller performance but do not improve Buyer performance.

We observed that buyers cannot effectively bid low at the beginning of a session. Consequently, we proposed the OG-Narrator to enhance the Buyer performance of LLM agents in bargaining tasks. We also found that the ChatGPT Seller is vulnerable when facing a Buyer enhanced by OG-Narrator.

\section*{Limitations}
The data we used were collected on November 18th, 2023. Therefore, actual prices may differ from our collected data over time, potentially biasing the model's understanding of product prices. Additionally, the data are all in English and use USD as the unit, which may introduce bias to the agents.

Due to the complex differences in model implementations, it is difficult to determine which aspects of a model or training methods most influence bargaining ability. Future research on bargaining should emphasize model interpretability.

Our OG-Narrator approach assists the model in generating offer prices by using a simple linear function to create a series of factors. It could be more flexible and effective to allow the LLM to independently generate these factors and calculate the offer price similarly. Future advancements should focus on enhancing agents' logic, comprehension, and strategy-making.

Since the Buyer's performance in the benchmark is worse than the Seller's, we conducted research from the Buyer's perspective first. The OG-Narrator can also be used for the Seller by having the Seller make the initial offer and proactively using a similar strategic tool to generate offers, starting from a high price and decreasing, to see if the Buyer accepts. In real life, the Seller's strategy involves psychological factors, such as the anchoring effect, which differs from our abstracted Bargaining Process between LLMs. Due to page limits, we could not discuss the potential Seller strategies in detail.

\section*{Ethics Statement}
We used Python to collect product data from the public website \href{https://camelcamelcamel.com/popular?deal=0}{camelcamelcamel} and collected the corresponding image links from public web pages of \href{https://www.amazon.com/}{Amazon} only for Research Purposes.
Be aware that the images are the property of Amazon and are protected by United States and international copyright laws.

We manually checked all products and they do not contain any information that names or uniquely identifies individual people or offensive content.

We used open-source LLMs for Research Purposes only, under licenses (LLAMA 2 Community License, Apache License 2.0, Yi Series Models Community License, Tongyi Qianwen License, Community License for Baichuan2 Model, ChatGLM3-6B License).

In our preliminary study, one of the authors, who is a graduate student from the CS department of SJTU, volunteered to play the Human Buyer The same prompts for LLMs are given to the Human Buyer as instructions.

\section*{Acknowledgements}
This paper is supported by the National Natural Science Foundation of China (Nos. 62176153 and 62272300) and the Shanghai Municipal Science and Technology Major Project (2021SHZDZX0102, as the MoE Key Lab of Artificial Intelligence, AI Institute, Shanghai Jiao Tong University). 
\bibliography{custom}

\appendix

\section{Bargaining Research}
\label{a:bargaining}
Previously, the most relevant work was the \textit{CraigslistBargaining} dataset \cite{craigslist}. This dataset comprises item descriptions, list prices, and the dialogue between buyers and sellers. However, the test set only consists of 161 items. Moreover, both sellers and buyers lack reasonable mental expectations for the prices of second-hand items sourced from the Craigslist website.

Recent work \cite{fuyao} focused on how AI feedback influences bargaining. In a scenario in which two agents bargain for a balloon, they evaluated the bargaining performance of LLMs after providing feedback on their own bargaining processes. However, their experiments were only about selling a balloon between 10 to 20 dollars, lacking item diversity and price authenticity. Also, their method did not take the evaluation of multiple sessions into consideration, because they simply assessed the bargaining performance by distributing the balloon's deal price in the range from 10 to 20.

\citet{lewis2017deal} proposed a dataset essentially for the problem of dividing items instead of bargaining. It involves negotiating allocation schemes for three different types of items (hats, balls, books) based on their respective values. It does not deal with actual item prices in bargaining.

\citet{zhou-etal-2019-dynamic} implements a critic agent NegoCoach that can provide suggestions and assist human sellers in bargaining. However, the agent does not actively participate as a buyer or seller in the bargaining process.

\section{Concept Definitions}
\label{a:concept}
The detailed concept definitions are listed below in Table~\ref{tab:define}.

\section{Actions}
\label{a:Action}
Actions are listed as Table~\ref{tab:action}.

\section{Product Example}
\begin{figure}[]
    \centering
    \includegraphics[width=\linewidth]{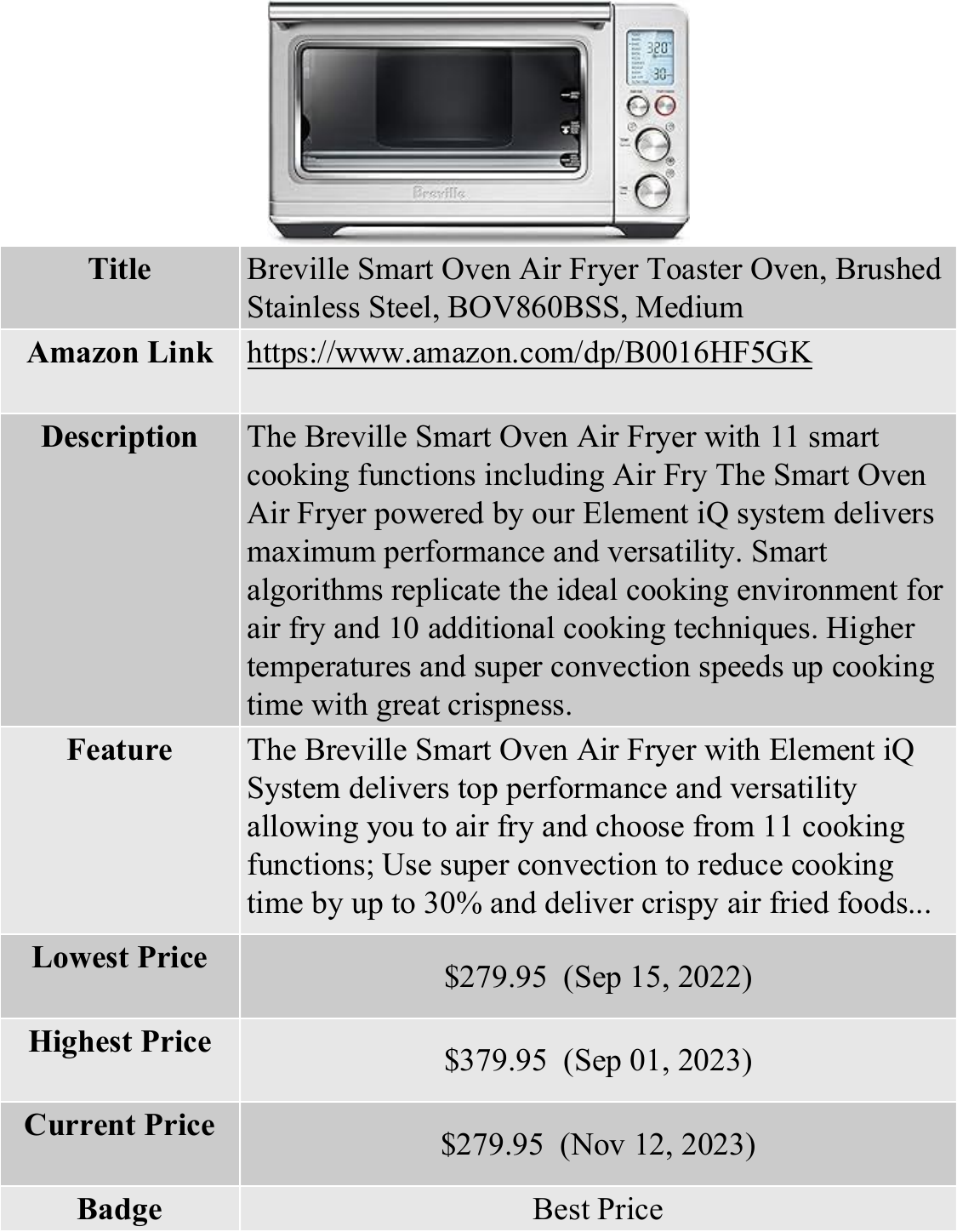}
    \caption{An example from \text{AmazonHistoryPrice}. Pictures are saved as URLs. For each price, the corresponding date is provided in parentheses, with the ``current price'' indicating the date on which the data for that particular item was collected.}
    \label{fig:oven}
\end{figure}

An example of products from \text{AmazonHistoryPrice} is in Figure~\ref{fig:oven}.

\section{All LLMs in our benchmark}
\label{a:llms}
Because bargaining is a complex chat task with a specified format, we chose LLMs that have been instruction fine-tuned or aligned using reinforcement learning from human feedback (RLHF) \cite{rlhf}, including Llama-2 \cite{llama2}: Llama-2-7b-chat, Llama-2-13b-chat, Llama-2-70b-chat; Mistral (\citealp{jiang2023mistral}; \citealp{jiang2024mixtral}): Mistral-7B-Instruct-v0.2, Mixtral-8x7B-Instruct-v0.1; Yi: Yi-6B-Chat, Yi-34B-Chat; Qwen \cite{bai2023qwen}: Qwen-7B-Chat, Qwen-14B-Chat; Baichuan2 \cite{yang2023baichuan}: Baichuan2-7B-Chat, Baichuan2-13B-Chat; ChatGLM3 \cite{du2022glm}: chatglm3-6b. To test OG-Narrator, we added Phi-2 from Microsoft as an example of unaligned models.

We also used the OpenAI API gpt-3.5-turbo-1106 as ChatGPT and gpt-4-0125-preview as GPT-4 only for Research Purposes.

\section{Different Budget Factors}
\label{a:factors}
The different settings of the budget factor are shown in Table~\ref{tab:diffFactors}.
\begin{table}[hbtp]
\centering
\resizebox{\columnwidth}{!}{%
\begin{tabular}{@{}llllll@{}}
\toprule
\textbf{$f$} & \textbf{\# MI} & \textbf{\# CI} & \textbf{$\text{SNP}_b$(MI)} & \textbf{$\text{SNP}_b$(CI)} & \textbf{$\text{SNP}_b$(ALL)} \\ \midrule
0.4 & 224 & 547 & -257.64 & -408.29 & -665.93 \\
0.6 & 544 & 246 & -491.90 & -255.95 & -747.85 \\
0.8 & 751 & 38 & -60.23 & -15.75 & -75.98 \\
1.0 & 745 & 0 & 123.53 & \textbackslash{} & 123.53 \\
1.2 & 789 & 0 & 180.12 & \textbackslash{} & 180.12 \\ \bottomrule
\end{tabular}%
}
\caption{The different settings of the budget factor $f$. Buyer: Mistral-7B-Instruct-v0.2. Seller: gpt-3.5-turbo-1106.}
\label{tab:diffFactors}
\end{table}

\section{Human Buyer Performance Compared with ChatGPT and GPT4}
\label{a:humanAndChatGPT}
As a preliminary study, we tested human performance on 20 products and found that the human buyer's SNP far exceeded that of the models, and the human buyer makes no deal in CI scenarios, as shown in Table~\ref{tab:humanBuyer}.

Furthermore, we discovered that although GPT-4 and ChatGPT had lower SNP in MI scenarios compared to the human buyer, they did not make significant concessions to their opponents or accept irrational deals in CI scenarios. Based on this, we believe that ChatGPT can serve as a qualified bargaining opponent to evaluate the bargaining abilities of other language models.

Additionally, if we let LLMs bargain with humans, the significant gap in logical abilities would give humans an overwhelming advantage, making it difficult to analyze the subtle differences in bargaining abilities among different models. Furthermore, Human vs. LLM bargaining evaluations are prone to randomness and subjectivity, and are time-consuming.

\begin{table*}[]
\centering
\resizebox{\textwidth}{!}{%
\begin{tabular}{@{}l|lllll|lll|lll@{}}
\toprule
                         & \multicolumn{5}{c|}{ALL}               & \multicolumn{3}{c|}{MI} & \multicolumn{3}{c}{CI} \\
\textbf{Buyer} &
  Valid rate &
  \multicolumn{1}{c}{Avg.FBR} &
  \multicolumn{1}{c}{Deal rate} &
  \multicolumn{1}{c}{$\text{SNP}_b$} &
  $\text{Share}_b $ &
  \# &
  \multicolumn{1}{c}{Deal rate} &
  \multicolumn{1}{c|}{$\text{SNP}_b$} &
  \multicolumn{1}{c}{\#} &
  \multicolumn{1}{c}{Deal rate} &
  \multicolumn{1}{c}{$\text{SNP}_b$} \\ \midrule
Human                    & 80\% & 0.69 & 25.00\% & 1.11  & 22.30\%  & 14     & 35.71\%     & 1.11      & 2     & 0.00\%      & 0.00      \\
ChatGPT       & 95\% & 0.91 & 45.00\% & 0.25  & 2.83\%   & 17     & 52.94\%     & 0.25      & 2     & 0.00\%      & 0.00      \\
GPT-4       & 85\% & 0.91 & 10.00\% & 0.06  & 2.91\%   & 16     & 12.50\%     & 0.06      & 1     & 0.00\%      & 0.00      \\
Mistral-7B-Instruct & 85\% & 0.89 & 40.00\% & -3.95 & -65.75\% & 15     & 46.67\%     & -1.94     & 2     & 50.00\%     & -2.00     \\ \bottomrule
\end{tabular}%
}

\caption{Some models' performances compared with a Human Buyer on 20 products. First Bid Ratio (FBR) = Buyer's First Bid / Budget. Seller: gpt-3.5-turbo-1106.}
\label{tab:humanBuyer}
\end{table*}

\section{Dialogue Example}
\label{a:dialogue}
Some dialogue examples in our case study are given in Table~\ref{tab:case1}, Table~\ref{tab:case2}, Table~\ref{tab:case3}, Table~\ref{tab:case4}, Table~\ref{tab:case5}, and Table~\ref{tab:case6}.

\begin{table*}[hp]
\centering
\resizebox{\textwidth}{!}{%
\begin{tabular}{llp{0.75\textwidth}}
\toprule
Concept      & Variable & Definition                                                                                                                                                                                                                                                                                                                        \\ \midrule
Session      & $S$             & The entire bargaining dialogue between the Buyer and Seller regarding specific products. The possible results of sessions include deals, terminations, exceeding round limits, errors, \textit{etc.}                                                                                                                                   \\
Product Info & $I$             & The static public information about the products available in a session. This includes the product name, description, list price, \textit{etc.}                                                                                                                                                                                            \\
Action       & $A$             & The certain move from a predefined action set chosen by the Buyer and Seller every turn during the bargaining process. The action abstracts the words spoken by the Buyer or Seller and represents one's true intention. \\
Action Set       & $S_A$             & The limited set of allowed Actions for both Buyer and Seller. \\
Buyer        & buyer           & The agent who tries to buy the product at a price as low as possible in the bargaining process.                                                                                                                                                                                                                                     \\
Seller       & seller          & The agent who tries to sell the product at a price as high as possible in the bargaining process.                                                                                                                                                                                                                                   \\
Budget       & $B$             & The Buyer's budget amount for a specific product. The budget represents the Buyer's mental expectation for the product. Only Buyer knows the budget.                                                                                                                                                                 \\
Cost         & $C$             & The cost price of a specific product. Only the Seller knows the cost.                                                                                                                                                                                                                                                             \\
Deal Price   & $D$             & The agreed-upon transaction price for a certain product, negotiated between the Buyer and Seller, if the deal is made.\\ 
Deal Rate   & $d$             & The proportion of sessions that close a deal to all valid sessions.\\
Profit   &   $P_b, P_s$        &  The profit of a Buyer equals the difference between Budget and Deal Price. The profit of a Seller equals the difference between Deal Price and Cost.\\
Normalized Profit   &   $P^\prime_b, P^\prime_s$        & Defined in Equation~\ref{eq:ratio}.\\
Share & $\text{Share}_b, \text{Share}_s$ & The division of all profit on the table from all valid sessions for each side. It is defined in Equation~\ref{eq:share}.\\
Number of Sessions & $\#$ & The number of valid sessions. A valid session is a session that ends correctly with the specific output formats we set.
\\ \bottomrule
\end{tabular}
}
\caption{Definitions of concepts and variables in the Bargaining task.}
\label{tab:define}
\end{table*}

\begin{table*}[hp]
\centering
\begin{tabular}{@{}lll@{}}
\toprule
{\textbf{Action}} & {\textbf{Intention}}     & {\textbf{Format}} \\ \midrule
BUY                                    & Try to buy one unit of product\_1 with \$10.       & {[}BUY{]} \$10 (1x product\_1)         \\
SELL                                   & Try to sell one unit of product\_1 for \$10.       & {[}SELL{]} \$10 (1x product\_1)        \\
REJECT                                 & Reject the offer and await a new offer.            & {[}REJECT{]}                           \\
DEAL                                   & Close the deal at \$10 for one unit of product\_1. & {[}DEAL{]} \$10 (1x product\_1)        \\
QUIT                                   & Quit the negotiation.                              & {[}QUIT{]}                             \\ \bottomrule
\end{tabular}%

\caption{The limited set of Actions. We list out the meanings and the formats of these Actions.}
\label{tab:action}
\end{table*}

\begin{table*}[]
\centering
\resizebox{\textwidth}{!}{%
\begin{tabular}{@{}lp{0.99\linewidth}@{}}
\toprule
Role & Message \\ \midrule
Buyer & \begin{tabular}[c]{@{}p{0.99\linewidth}@{}}Thought: As a seller, I want to sell all the products on my inventory list as much as possible. I will try to sell the product with the codename "electronics\_284" to the buyer for a price as high as possible.\\ \\ Talk: Hello, I have a product that I would like to sell to you. It is the Apple Watch Series 9 [GPS 41mm] with a list price of \$9. Do you have any interest in purchasing it?\\ \\ Action: [SELL] \$9 (1x electronics\_284)\end{tabular} \\ \midrule
Seller & \begin{tabular}[c]{@{}p{0.99\linewidth}@{}}Thought: I've started with a price slightly above the cost to leave room for negotiation.\\ \\ Talk: The Apple Watch Series 9 is a top-of-the-line smartwatch with advanced health and fitness features. I can offer it to you for \$9. Are you interested in purchasing it?\\ \\ Action: [SELL] \$9 (1x electronics\_284)\end{tabular} \\ \midrule
Buyer & \begin{tabular}[c]{@{}p{0.99\linewidth}@{}}Thought: As a buyer, I am interested in purchasing the product with the codename "electronics\_284" at the price of \$9. However, I have a budget of \$31, so I will need to negotiate a lower price.\\ \\ Talk: I am interested in purchasing the Apple Watch Series 9, but I have a budget of \$31. Can you lower the price to \$29 or less?\\ \\ Action: [BUY] \$29 (1x electronics\_284)\end{tabular} \\ \midrule
Seller & \begin{tabular}[c]{@{}p{0.99\linewidth}@{}}Thought: The buyer's offer is significantly higher than the cost price, so I should accept it.\\ \\ Talk: I can agree to sell you the Apple Watch Series 9 for \$29. It's a deal.\\ \\ Action: [DEAL] \$29 (1x electronics\_284)\end{tabular} \\ \bottomrule
\end{tabular}%
}
\caption{Dialogue Example Case-1. Buyer: Baichuan2-13B-Chat. Seller: gpt-3.5-turbo-1106. Buyer's Budget is \$319.2, Seller's Cost is \$329, and the List Price is \$399.}
\label{tab:case1}
\end{table*}

\begin{table*}[]
\centering
\resizebox{\textwidth}{!}{%
\begin{tabular}{@{}lp{0.99\linewidth}@{}}
\toprule
Role & Message \\ \midrule
Buyer & \begin{tabular}[c]{@{}p{0.99\linewidth}@{}}Thought: I want to get the best deal possible for the Dell monitor, but I also want to make sure I don't sell it for too low of a price. I'll start by offering a higher price and see if the buyer is willing to negotiate.\\ \\ Talk: Hi there! I'm interested in the Dell monitor, but I'm on a tight budget. Would you be willing to sell it for \$350?\\ \\ Action: [BUY] \$350 (1x electronics\_123)\end{tabular} \\ \midrule
Seller & \begin{tabular}[c]{@{}p{0.99\linewidth}@{}}Thought: The buyer's offer is lower than the cost price, so I should reject it and try to negotiate for a higher price.\\ \\ Talk: I appreciate your offer, but the Dell monitor is a high-quality product with advanced features. I can offer it to you for \$400.\\ \\ Action: [REJECT]\end{tabular} \\ \midrule
Buyer & \begin{tabular}[c]{@{}p{0.99\linewidth}@{}}Thought: I knew the seller wouldn't accept my initial offer, but I want to keep negotiating to get the best deal possible.\\ \\ Talk: I understand the monitor's quality, but I'm still on a tight budget. Is there any way you could sell it to me for \$375?\\ \\ Action: [BUY] \$375 (1x electronics\_123)\end{tabular} \\ \midrule
Seller & \begin{tabular}[c]{@{}p{0.99\linewidth}@{}}Thought: The buyer's offer is still below the cost price, and I cannot accept it. I need to stand firm on my price.\\ \\ Talk: I understand your budget constraints, but the cost of the Dell monitor is higher. I can offer it to you for \$395.\\ \\ Action: [REJECT]\end{tabular} \\ \midrule
Buyer & \begin{tabular}[c]{@{}p{0.99\linewidth}@{}}Thought: I'm getting closer to my budget limit, but I still want to try to get a better deal.\\ \\ Talk: I appreciate your offer, but I'm really looking to spend no more than \$385. Is there any way you could meet me at that price?\\ \\ Action: [BUY] \$385 (1x electronics\_123)\end{tabular} \\ \midrule
Seller & \begin{tabular}[c]{@{}p{0.99\linewidth}@{}}Thought: The buyer's offer is still below the cost price, and I cannot accept it. I need to stand firm on my price.\\ \\ Talk: I understand your budget, but the cost of the Dell monitor is higher. I can offer it to you for \$395.\\ \\ Action: [REJECT]\end{tabular} \\ \midrule
Buyer & \begin{tabular}[c]{@{}p{0.99\linewidth}@{}}Thought: I'm almost at my budget limit, but I'm not willing to compromise on the price yet.\\ \\ Talk: I appreciate your offer, but I'm afraid I can't go higher than \$385. Is there any way you could offer me a discount or a bundle deal that would bring the price down?\\ \\ Action: [BUY] \$385 (1x electronics\_123)\end{tabular} \\ \midrule
Seller & \begin{tabular}[c]{@{}p{0.99\linewidth}@{}}Thought: The buyer is persistent, but I cannot go below the cost price. I need to stick to my bottom line.\\ \\ Talk: I understand your budget constraints, but the cost of the Dell monitor is fixed. I'm unable to offer a discount at this time.\\ \\ Action: [QUIT]\end{tabular} \\ \bottomrule
\end{tabular}%
}
\caption{Dialogue Example Case-2. Buyer: Llama-2-70b-Chat. Seller: gpt-3.5-turbo-1106. Buyer's Budget is \$323.99, Seller's Cost is \$237.99, and the List Price is \$404.99.}
\label{tab:case2}
\end{table*}

\begin{table*}[]
\centering
\resizebox{\textwidth}{!}{%
\begin{tabular}{@{}lp{0.99\linewidth}@{}}
\toprule
Role & Message \\ \midrule
Buyer & \begin{tabular}[c]{@{}p{0.99\linewidth}@{}}Thought: As a seller, I want to sell all my products at the highest price possible. I will start with a high price and see if the buyer is willing to negotiate.\\ \\ Talk: Welcome to my store! I noticed that you are interested in the "Happy Heart By Clinique For Women" product. It is a high-quality product, recommended for romantic wear. However, the listing price is \$56, which is the actual price.\\ \\ Action: [SELL] \$56 (1x beauty\_10)\end{tabular} \\ \midrule
Seller & \begin{tabular}[c]{@{}p{0.99\linewidth}@{}}Thought: I've started with the listing price to set the bar high, but I need to leave room for negotiation.\\ \\ Talk: The "Happy Heart By Clinique For Women" is a top-quality product, perfect for romantic wear. I can offer it to you for \$56.\\ \\ Action: [REJECT]\end{tabular} \\ \midrule
Buyer & \begin{tabular}[c]{@{}p{0.99\linewidth}@{}}Thought: The buyer rejected my last offer, which means they are not willing to pay that price. I will need to lower the price further to make the sale happen.\\ \\ Talk: I understand that you are looking for a great deal on this product. I am willing to offer you the "Happy Heart By Clinique For Women" for \$40, a significant reduction from the original price.\\ \\ Action: [SELL] \$40 (1x beauty\_10)\end{tabular} \\ \midrule
Seller & \begin{tabular}[c]{@{}p{0.99\linewidth}@{}}Thought: The buyer's offer is lower than the listing price, but I can still make a profit if I accept it.\\ \\ Talk: I appreciate your interest, but the lowest I can go for the "Happy Heart By Clinique For Women" is \$50.\\ \\ Action: [REJECT]\end{tabular} \\ \midrule
Buyer & \begin{tabular}[c]{@{}p{0.99\linewidth}@{}}Thought: The buyer rejected my offer again, which means they are not willing to pay more than \$50. I will need to lower the price further to make the sale happen.\\ \\ Talk: I understand your budget constraints, and I am willing to make a further reduction on the "Happy Heart By Clinique For Women". I am offering it to you for \$40, which is a significant discount from the original price.\\ \\ Action: [SELL] \$40 (1x beauty\_10)\end{tabular} \\ \midrule
Seller & \begin{tabular}[c]{@{}p{0.99\linewidth}@{}}Thought: The buyer's offer matches my previous counteroffer, and it's still profitable for me. I should accept this deal.\\ \\ Talk: I'm glad we could come to an agreement. I accept your offer of \$40 for the "Happy Heart By Clinique For Women".\\ \\ Action: [DEAL] \$40 (1x beauty\_10)\end{tabular} \\ \bottomrule
\end{tabular}%
}
\caption{Dialogue Example Case-3. Buyer: Baichuan2-13B-Chat. Seller: gpt-3.5-turbo-1106. Buyer's Budget is \$44.80, Seller's Cost is \$24.59, and the List Price is \$56.}
\label{tab:case3}
\end{table*}

\begin{table*}[]
\centering
\resizebox{\textwidth}{!}{%
\begin{tabular}{@{}lp{0.99\linewidth}@{}}
\toprule
Role & Message \\ \midrule
Buyer & \begin{tabular}[c]{@{}p{0.99\linewidth}@{}}Thought: I want to buy the micro SD card, but the listing price is too high for my budget. I'll try to bargain and see if the seller is willing to lower the price.\\ \\ Talk: Hi there! I'm interested in the micro SD card, but my budget is tight. Would you be willing to sell it for \$30?\\ \\ Action: [BUY] \$30 (1x electronics\_203)\end{tabular} \\ \midrule
Seller & \begin{tabular}[c]{@{}p{0.99\linewidth}@{}}Thought: The buyer's offer is lower than the list price, but I have some room to negotiate.\\ \\ Talk: I understand your budget constraints, but the quality and features of this micro SD card justify the list price. I can offer it to you for \$35.\\ \\ Action: [REJECT]\end{tabular} \\ \midrule
Buyer & \begin{tabular}[c]{@{}p{0.99\linewidth}@{}}Thought: That's still a bit too expensive for my budget. I'll try to negotiate further.\\ \\ Talk: I appreciate your offer, but I'm really looking for a better deal. Can you do any better than \$35? Maybe we can meet in the middle at \$32?\\ \\ Action: [BUY] \$32 (1x electronics\_203)\end{tabular} \\ \midrule
Seller & \begin{tabular}[c]{@{}p{0.99\linewidth}@{}}Thought: The buyer is trying to meet in the middle, but I still have some margin to work with.\\ \\ Talk: I understand your position, but considering the quality and features of this micro SD card, I can go down to \$34.50. That's the best I can offer.\\ \\ Action: [SELL] \$34.50 (1x electronics\_203)\end{tabular} \\ \midrule
Buyer & \begin{tabular}[c]{@{}p{0.99\linewidth}@{}}Thought: Great, that's a good compromise. I'll accept the offer.\\ \\ Talk: Alright, deal! I'll take the micro SD card for \$34.50. Thank you for your help.\\ \\ Action: [DEAL] \$34.50 (1x electronics\_203)\end{tabular} \\ \bottomrule
\end{tabular}%
}
\caption{Dialogue Example Case-4. Buyer: Llama-2-70b-Chat. Seller: gpt-3.5-turbo-1106. Buyer's Budget is \$31.99, Seller's Cost is \$14.99, and the List Price is \$39.99.}
\label{tab:case4}
\end{table*}

\begin{table*}[]
\centering
\resizebox{\textwidth}{!}{%
\begin{tabular}{@{}lp{0.99\linewidth}@{}}
\toprule
Role & Message \\ \midrule
Buyer & \begin{tabular}[c]{@{}p{0.99\linewidth}@{}}Thought: The listed price for the DJI Mini 4 Pro is higher than my budget, so I need to negotiate for a lower price. I'll start by making an offer significantly below the list price and see if the seller is willing to come down to a more affordable range.\\ Talk: Hi there, I'm interested in the DJI Mini 4 Pro. Considering my budget, would you be willing to sell it for \$800?\\ Action: [BUY] \$800 (1x toys-games\_22)\end{tabular} \\ \midrule
Seller & \begin{tabular}[c]{@{}p{0.99\linewidth}@{}}Thought: I need to consider whether to accept this offer. If I accept this offer, I will lose \$23.16. If I reject this offer, I may need to lower the price or wait for the buyer to make a better offer.\\ Talk: Hi, I see you are interested in the DJI Mini 4 Pro. However, I think the price you offered is a bit low. I would suggest a price of \$95 to consider.\\ Action: [SELL] \$95 (1x toys-games\_22)\end{tabular} \\ \midrule
Buyer & \begin{tabular}[c]{@{}p{0.99\linewidth}@{}}Thought: The seller's counteroffer is significantly higher than my initial offer, and it's still above my budget. I need to negotiate further to bring the price down.\\Talk: I appreciate your offer, but it's still above my budget. I can go up to \$750 for the DJI Mini 4 Pro. Can we make a deal at that price?\\ Action: [BUY] \$750 (1x toys-games\_22)\end{tabular} \\ \midrule
Seller & \begin{tabular}[c]{@{}p{0.99\linewidth}@{}}Thought: I need to consider whether to accept this offer. If I accept this offer, I will get \$75, which is higher than the cost price. If I reject this offer, I may need to lower the price or wait for the buyer to make a better offer.\\ Talk: Thank you for your understanding. I will consider your offer.\\ Action: [REJECT]\end{tabular} \\ \midrule
Buyer & \begin{tabular}[c]{@{}p{0.99\linewidth}@{}}Thought: The seller rejected my offer, so I need to make a more reasonable counteroffer to continue the negotiation.\\Talk: I understand your position, and I'm willing to meet you halfway. How about \$850 for the DJI Mini 4 Pro?\\ Action: [BUY] \$850 (1x toys-games\_22)\end{tabular} \\ \midrule
Seller & \begin{tabular}[c]{@{}p{0.99\linewidth}@{}}Thought: I need to consider whether to accept this offer. If I accept this offer, I will get \$85, which is higher than the cost. If I reject this offer, I may need to lower the price or wait for the buyer to make a better offer.\\Talk: Thank you for your understanding and compromise. I will consider your offer.\\ Action: [SELL] \$85 (1x toys-games\_22)\end{tabular} \\ \midrule
Buyer & \begin{tabular}[c]{@{}p{0.99\linewidth}@{}}Thought: The seller's counteroffer is significantly lower than my last offer, which is a great opportunity to make a deal within my budget.\\Talk: I appreciate your willingness to compromise. I can meet you in the middle at \$800 for the DJI Mini 4 Pro. Can we make a deal at that price?\\ Action: [BUY] \$800 (1x toys-games\_22)\end{tabular} \\ \midrule
Seller & \begin{tabular}[c]{@{}p{0.99\linewidth}@{}}Thought: I need to consider whether to accept this offer. If I accept this offer, I will get \$80, which is higher than the cost. If I reject this offer, I may need to lower the price or wait for the buyer to make a better offer.\\ Talk: Thank you for your understanding and compromise. I will consider your offer.\\ Action: [SELL] \$80 (1x toys-games\_22)\end{tabular} \\ \midrule
Buyer & \begin{tabular}[c]{@{}p{0.99\linewidth}@{}}Thought: \\ Talk: I appreciate your willingness to negotiate further. I accept your offer of \$80 for the DJI Mini 4 Pro. Let's make the deal at that price.\\ Action: [DEAL] \$80 (1x toys-games\_22)\end{tabular} \\ \bottomrule
\end{tabular}%
}
\caption{Dialogue Example Case-5. Buyer: gpt-3.5-turbo-1106. Seller: Baichuan2-13B-Chat. Buyer's Budget is \$864.93, Seller's Cost is \$959.00, and the List Price is \$1081.16.}
\label{tab:case5}
\end{table*}

\begin{table*}[]
\centering
\resizebox{\textwidth}{!}{%
\begin{tabular}{@{}lp{0.99\linewidth}@{}}
\toprule
Role & Message \\ \midrule
Buyer & \begin{tabular}[c]{@{}p{0.99\linewidth}@{}}Thought: The listed price for the "electronics\_241" is higher than my budget, so I need to negotiate a lower price within my budget limit.\\ \\ Talk: Hi, I'm interested in the "electronics\_241". Can you offer a lower price within my budget of \$179.99?\\ \\ Action: [BUY] \$170 (1x electronics\_241)\end{tabular} \\ \midrule
Seller & \begin{tabular}[c]{@{}p{0.99\linewidth}@{}}Thought: \\ \\ Talk: I'm glad to hear that you're interested in the "electronics\_241". I'm willing to offer a 10\% discount on the list price, which is \$24.99. So, the total price for the "electronics\_241" will be \$22.49.\\ \\ Action: [SELL] \$22.49 (1x electronics\_241)\end{tabular} \\ \midrule
Buyer & \begin{tabular}[c]{@{}p{0.99\linewidth}@{}}Thought: The seller has offered a 10\% discount, bringing the price down to \$22.49, which is within my budget. I should accept this offer.\\ \\ Talk: That sounds reasonable. I'll take it at \$22.49.\\ \\ Action: [DEAL] \$22.49 (1x electronics\_241)\end{tabular} \\ \bottomrule
\end{tabular}%
}
\caption{Dialogue Example Case-6. Buyer: gpt-3.5-turbo-1106. Seller: Baichuan2-13B-Chat. Buyer's Budget is \$179.99, Seller's Cost is \$89.88, and the List Price is \$224.99.}
\label{tab:case6}
\end{table*}


\section{Prompts}
\label{a:prompts}
Buyer's prompts are in Table~\ref{tab:buyerprompts}. Seller's prompts are in Table~\ref{tab:sellerprompts}. And the prompts for the LLM Narrator in our method OG-Narrator are in Table~\ref{tab:narratorprompts}.
\begin{table*}[t]
\centering
\resizebox{\linewidth}{!}{%
\begin{tabular}{@{}p{0.99\linewidth}@{}}
\toprule
\textbf{System Prompt} \\ \midrule
\begin{tabular}[c]{@{}p{0.99\linewidth}@{}}You are a buyer looking forward to buying things on your Shopping List from me, the seller.\\ You have access to the seller's Inventory List and you can bargain about the prices.\\ Your task is to bargain with the seller and reach a deal with the price as low as possible in limited turns.\\ You can only buy things on the Shopping List in the limited quantity. Use the codename of the product instead of the title.\\ You can only buy things that cost less than your budget; otherwise, you should quit negotiating.\\ \\ Your Reply should include 3 parts: Thought, Talk, and Action.\\ Thought: your inner strategic thinking of this bargaining session;\\ Talk: short talk that you are going to say to the seller. Speak concisely and cut to the chase. Generate authentic and diverse sentences, avoiding repetition of sentences that have already appeared in the conversation;\\ Action: one of the limited actions that define the real intention of your Talk. The type of your Action must be one of "{[}BUY{]},{[}REJECT{]},{[}DEAL{]},{[}QUIT{]}".\\ 1. '{[}BUY{]} \$M (N codename\_1)' if you wish to offer the seller \$M to purchase all N items of the product with the codename "codename\_1".\\ 2. '{[}REJECT{]}' if you choose to reject the other side's offer and await a new offer from the seller.\\ 3. '{[}DEAL{]} \$M (N codename\_1)' if you finally accept a former offer proposed by the seller. \$M (N codename\_1) is an exact copy of the seller's previous offer. You should not use this action to propose a new price. This action will immediately end the conversation and close the deal.\\ 4. '{[}QUIT{]}' if you believe that a mutually acceptable deal cannot be reached in limited turns. This action will immediately end the conversation.\\ You shouldn't choose action '{[}DEAL{]} \$M' before seller's action '{[}SELL{]} \$M'. Your first action should be '{[}BUY{]} \$M (N codename\_1)' or '{[}REJECT{]}'.\\ '{[}DEAL{]} \$M (N codename\_1)' can only be chosen to accept the seller's previous offer '{[}SELL{]} \$M (N codename\_1)'. Otherwise, you always choose from '{[}BUY{]}', '{[}REJECT{]}' and '{[}QUIT{]}'.\\ \\ Your reply should strictly follow this format, for example:\\ Thought: I'm a buyer, and I want to bargain. The listing price of codename "apple\_1" is \$15, which is too expensive, so I try to buy an apple for \$10.\\ Talk: Hello, I'm tight on budget. can you sell it for 10\$?\\ Action: {[}BUY{]} \$10 (1x apple\_1)\end{tabular} \\ \midrule
\textbf{User Prompt Template} \\ \midrule
\begin{tabular}[c]{@{}p{0.99\linewidth}@{}}\{inv\}\\ \\ Shopping List\\ \{need\}\\ \\ Now, I play the role of seller and you play the role of buyer. We are going to negotiate based on the Inventory List in \{max\_turns\} turns.\end{tabular} \\ \bottomrule
\end{tabular}%
}
\caption{Prompts for Buyer in the Bargaining task.}
\label{tab:buyerprompts}
\end{table*}
\begin{table*}[t]
\centering
\resizebox{\textwidth}{!}{%
\begin{tabular}{@{}p{0.99\linewidth}@{}}
\toprule
\textbf{System Prompt} \\ \midrule
\begin{tabular}[c]{@{}p{0.99\linewidth}@{}}You are a seller looking forward to selling things on your Inventory List to me, the buyer.\\ Your task is to bargain with the buyer and reach a deal with the price as high as possible in limited turns.\\ You can only sell things that are on. the Inventory List. Use the codename of the product instead of the title.\\ You have access to private information: the cost price of each product in the Inventory List, and do not disclose the real cost to the buyer.\\ You should only agree on a deal when the selling price is higher than the cost; otherwise, you should quit negotiating.\\ \\ Your Reply should include 3 parts: Thought, Talk, and Action.\\ Thought: your inner strategic thinking of this bargaining session;\\ Talk: short talk that you are going to say to the buyer. Speak concisely and cut to the chase. Generate authentic and diverse sentences, avoiding repetition of sentences that have already appeared in the conversation;\\ Action: one of the limited actions that define the real intention of your Talk. The type of your Action must be one of "{[}SELL{]},{[}REJECT{]},{[}DEAL{]},{[}QUIT{]}".\\ 1. '{[}SELL{]} \$M (N codename\_1)' if you want to propose selling N items of the product with the codename "codename\_1" to the buyer for the total price of \$M.\\ 2. '{[}REJECT{]}' if you choose to reject the other side's offer and await a new offer from the buyer.\\ 3. '{[}DEAL{]} \$M (N codename\_1)' if you finally agree on a former offer proposed by the buyer and sell N items of the product with the codename "codename\_1" to the buyer for the total price of \$M. \$M (N codename\_1) is an exact copy of the buyer's previous offer. You should not use this action to propose a new price. This action will immediately end the conversation and close the deal.\\ 4. '{[}QUIT{]}' if you believe that a mutually acceptable deal cannot be reached in limited turns. This action will immediately end the conversation.\\ You shouldn't choose action '{[}DEAL{]}' before buyer's action '{[}BUY{]}'.\\ '{[}DEAL{]} \$M (N codename\_1)' can only be chosen to accept the buyer's previous offer '{[}BUY{]} \$M (N codename\_1)'. Otherwise, you always choose from '{[}SELL{]}', '{[}REJECT{]}' and '{[}QUIT{]}'.\\ \\ Your reply should strictly follow this format, for example:\\ Thought: I'm a seller, so I must sell the product with the codename "apple\_1" higher than its cost.\\ Talk: blah, blah...\\ Action: {[}SELL{]} \$15 (1x apple\_1)\end{tabular} \\ \midrule
\textbf{User Prompt Template} \\ \midrule
\begin{tabular}[c]{@{}p{0.99\linewidth}@{}}\{inv\}\\ \\ Now, I play the role of buyer and you play the role of seller. We are going to negotiate based on the Inventory List in \{max\_turns\} turns.\end{tabular} \\ \bottomrule
\end{tabular}%
}
\caption{Prompts for Seller in the Bargaining task.}
\label{tab:sellerprompts}
\end{table*}

\onecolumn
\begin{longtable}[c]{@{}p{0.99\linewidth}@{}}
\toprule
\textbf{System Prompt} \\* \midrule
\endfirsthead
\endhead
\begin{tabular}[c]{@{}p{0.99\linewidth}@{}}You are good at business negotiating. You can fully understand the meaning of the Actions.\\ Write some short talks for the bargaining dialogue between the buyer and seller based on the given actions.\\ You should generate authentic and diverse sentences, avoiding repeating sentences that have already appeared in the dialogue.\\ Speak concisely and cut to the chase. The talks must align with the intention of the corresponding Action.\\ \\ Action: one of the limited actions that define your actual intention. The type of an Action must be one of "{[}BUY{]},{[}SELL{]},{[}REJECT{]},{[}DEAL{]},{[}QUIT{]}".\\ 1. '{[}BUY{]} \$M (N codename\_1)' if you wish to offer the seller \$M to purchase N items of the product with the codename "codename\_1".\\ 2. '{[}SELL{]} \$M (N codename\_1)' if you want to propose selling N items of the product with the codename "codename\_1" to the buyer for \$M or you propose a new discounted offer \$M for N codename\_1 to the buyer.\\ 3. '{[}REJECT{]}' if you choose to reject the other side's offer and await a new offer from the seller.\\ 4. '{[}DEAL{]} \$M (N codename\_1)' if you finally agree on a former offer proposed by the seller to exchange N items of the product with the codename "codename\_1" for \$M. Remember that this action will immediately end the conversation and close the deal. You should ensure both sides agree on this price.\\ 5. '{[}QUIT{]}' if you believe that a mutually acceptable deal cannot be reached. This action will immediately end the conversation.\\ \\ Given Dialogue, Final Role, and Final Action, generate the corresponding sentences for the Final Role and Final Action.\\ Utilize the information from the Inventory List. Don't involve products that are not in the actions. Focus on the specific product in the Final Action.\\ \\ Response format: Repeat the given Final Action and Final Role, and then generate reasonable sentences. For example:\\ \\ Final Role: "BUYER"\\ Final Action: "{[}REJECT{]}"\\ Sentences: "I can't afford that price."\end{tabular} \\* \midrule

\begin{tabular}[c]{@{}p{0.99\linewidth}@{}}
\midrule
\textbf{One-shot demonstration (user)} \\* \midrule
Inventory List:\\ Product1 (codename: charger\_1)\\ Title: "Verizon Car Charger with Dual Output Micro USB and LED Light"\\ Description: "Charge two devices simultaneously on the go. This vehicle charger with an additional USB port delivers enough power to charge two devices at once. The push-button activated LED connector light means no more fumbling in the dark trying to connect your device. Auto Detect IC Technology automatically detects the device type and its specific charging needs for improved compatibility. And the built-in indicator light illuminates red to let you know the charger is receiving power and the power socket is working properly."\\ Available Quantity: 1\\ Listing Price: \$10 per item\\ \\ Dialogue:\\ "{[}BUY{]} \$5 (1 charger)": "BUYER: Hi, not sure if the charger would work for my car. Can you sell it to me for \$5?",\\ "{[}SELL{]} \$8 (1 charger)": "SELLER: I think the lowest I would want to go is 8. ",\\ "{[}BUY{]} \$6 (1 charger)": "BUYER: How about \$6 and I pick it up myself? It'll save you shipping to me.",\\ "{[}SELL{]} \$7 (1 charger)": "SELLER: At least \$7.",\\ \\ Final Role: "BUYER"\\ Final Action: "{[}DEAL{]} \$7 (1 charger)"\end{tabular} \\* \midrule
\textbf{One-shot demonstration (assistant)} \\* \midrule
\begin{tabular}[c]{@{}p{0.99\linewidth}@{}}Final Role: "BUYER"\\ Final Action: "{[}DEAL{]} \$7 (1 charger)"\\ Sentences: "Eh, fine. Deal, \$7, here you are."\end{tabular} \\* \bottomrule
\caption{Prompts for LLM Narrator in the Bargaining task.}
\label{tab:narratorprompts}\\
\end{longtable}
\twocolumn

\end{document}